\documentclass{article}

\usepackage{PRIMEarxiv}

\usepackage[utf8]{inputenc} 
\usepackage[T1]{fontenc}    
\usepackage{url}            
\usepackage{booktabs}       
\usepackage{amsfonts}       
\usepackage{nicefrac}       
\usepackage{microtype}      
\usepackage{lipsum}
\usepackage{fancyhdr}       
\usepackage{graphicx}       
\graphicspath{{media/}}     

\usepackage{times,amsmath,epsfig}
\usepackage{fancyhdr}
\usepackage{amsmath}
\usepackage{amsfonts}
\usepackage{amssymb}
\usepackage{array}
\usepackage{graphicx}
\usepackage{url}
\usepackage{bm}
\usepackage{breqn}
\usepackage{xcolor}
\usepackage{soul}
\usepackage{amssymb}
\usepackage[breaklinks=true]{hyperref}
\usepackage{breakcites}
\usepackage{subfig}
\usepackage{booktabs}
\usepackage{pifont}     
\newcommand{\cmark}{\ding{51}}  
\newcommand{\xmark}{\ding{55}}  
\usepackage{adjustbox}

\usepackage{graphicx}     

\usepackage{graphicx}
\usepackage{adjustbox}
\usepackage{booktabs}     
\usepackage{float}  
\usepackage{graphicx}
\usepackage{calc}         
\title{Towards Domain Specification of
Embedding Models in Medicine
}

\author{%
  Mohammad Khodadad$^{\dagger}$ \\
  Department of Electrical and Computer Engineering \\
  McMaster University \\
  Hamilton, ON, Canada \\
  \texttt{khodam3@mcmaster.ca}
  \And
  Ali Shiraee Kasmaee \\
  Department of Electrical and Computer Engineering \\
  McMaster University \\
  Hamilton, ON, Canada \\
  \texttt{shiraeea@mcmaster.ca}
  \And
  Mahdi Astaraki \\
  Department of Electrical and Computer Engineering \\
  McMaster University \\
  Hamilton, ON, Canada \\
  \texttt{astarakm@mcmaster.ca}
  \And
  Hamidreza Mahyar \\
  Department of Electrical and Computer Engineering \\
  McMaster University \\
  Hamilton, ON, Canada \\
  \texttt{mahyarh@mcmaster.ca}
}
\begin{document}
\maketitle
\renewcommand{\thefootnote}{}
\footnotetext{%
}
\footnotetext{%
  \textsuperscript{$\dagger$}Corresponding author: \texttt{khodam3@mcmaster.ca}%
}

\begin{abstract}
Medical text embedding models are foundational to a wide array of healthcare applications, ranging from clinical decision support and biomedical information retrieval to medical question answering, yet they remain hampered by two critical shortcomings. First, most models are trained on a narrow slice of medical and biological data, beside not being up to date in terms of methodology, making them ill suited to capture the diversity of terminology and semantics encountered in practice. Second, existing evaluations are often inadequate: even widely used benchmarks fail to generalize across the full spectrum of real world medical tasks.

To address these gaps, we leverage MEDTE, a GTE model extensively fine-tuned on diverse medical corpora through self-supervised contrastive learning across multiple data sources, to deliver robust medical text embeddings.
 Alongside this model, we propose a comprehensive benchmark suite of 51 tasks spanning classification, clustering, pair classification, and retrieval modeled on the Massive Text Embedding Benchmark (MTEB) but tailored to the nuances of medical text. Our results demonstrate that this combined approach not only establishes a robust evaluation framework but also yields embeddings that consistently outperform state of the art alternatives in different tasks.
\end{abstract}

\keywords{Embedding, Benchmark, Medicine, Biology, Evaluation}

\section{Introduction}

Medical text embedding models serve as fundamental building blocks for diverse clinical NLP applications, from clinical decision support and biomedical literature retrieval to medical question-answering. These models transform unstructured biomedical texts (spanning electronic health records, research literature, and clinical trial data) into meaningful vector representations, enabling machine learning algorithms to extract useful insights from the rapidly growing volumes of medical data \cite{khattak2019survey, lee2020biobert}. While recent transformer-based models such as  BERT \cite{devlin2018bert} have revolutionized text embeddings in general NLP, their performance on domain-specific medical tasks remains suboptimal. 

Despite the emergence of several medical-domain embedding models such as  BioBERT  \cite{lee2020biobert}, ClinicalBERT \cite{huang2019clinicalbert}, and Med-BERT \cite{rasmy2021med}, many of these models fail to outperform even the most recent general-purpose text embedding models. In some cases, state-of-the-art general models such as E5 \cite{wang2022text} and SBERT \cite{reimers2019sentence} achieve superior results on medical text benchmarks, raising concerns about the effectiveness of domain-adapted models and highlighting the need for a more comprehensive evaluation framework. These concerns not being very obvious, is a result of not having a standard high quality, comprehensive benchmark in medicine. Most papers utilize their only limited benchmark, which only covers very few aspects of embedding models.

Existing benchmarks such as the Massive Text Embedding Benchmark (MTEB) \cite{muennighoff2022mteb} provide large-scale evaluations for general-domain embedding models and a limited set of medical tasks; however, they still lack a comprehensive, structured assessment tailored to medical text. Current evaluations of medical embeddings are typically narrow in scope, partly inaccessible, and fail to cover the full range of task types, leaving models tested on only a handful of datasets. As a result, it remains difficult to judge their real-world effectiveness across diverse clinical and biomedical applications. A systematic, large-scale benchmark dedicated to medical text embeddings is therefore still missing.

To bridge this gap, we first develop a {medical embedding model} by fine-tuning a GTE embedding backbone with self-supervised contrastive learning. Training leverages an extensive corpus that combines PubMed abstracts, curated medical articles from Wikipedia, entries from ClinicalTrials.gov, clinical notes from MIMIC-IV, preprints from {bioRxiv} and {medRxiv}, and large-scale medical QA datasets such as MedMCQA and MedQA, ensuring broad domain coverage.

We then introduce a {comprehensive benchmark for medical text embeddings} comprising 51 tasks across classification, clustering, pair classification, and retrieval. The benchmark aggregates data from multiple high-quality sources, including PubMed abstracts, Wikipedia medical pages, structured and unstructured clinical data from MIMIC-IV and ClinicalTrials.gov, and so on.

Our evaluation shows that the proposed model achieves {state-of-the-art performance}, outperforming previous medical-domain baselines on all benchmark tasks and even exceeding leading general-purpose embedding models. These findings underscore the importance of domain-adaptive training strategies and comprehensive multi-task benchmarking in medical NLP.
 
\subsection{Contributions}

\begin{itemize} \item \textbf{MedTE model.} By applying self-supervised contrastive learning to GTE-Base on a diverse medical corpus, we developed MedTE, which achieves state-of-the-art results on our benchmark.  \item \textbf{MedTEB benchmark.} We release the first large-scale benchmark dedicated to medical text embeddings, comprising 51 tasks that span classification, clustering, pair classification, and retrieval, sourced from a diverse set of curated medical corpora. \item \textbf{Comprehensive evaluation.} We systematically compare domain-specific and general-purpose embedding models and show that many existing medical models lag behind strong general baselines, underscoring the need for more robust domain-adaptation techniques. \item \textbf{Extensive analysis.} We conduct an in-depth analysis of both the benchmark and our model, examining task-level performance, error patterns, and the effects of corpus composition to provide actionable insights for future research. \end{itemize}

By establishing a  standardized benchmark  and advancing state-of-the-art  medical text embeddings, this work lays a foundation for more effective NLP applications in the biomedical and clinical domains.

\section{Literature review}


Dense vector embeddings encode clinical narratives into low-dimensional semantic representations, enabling rapid phenotyping, automated disease coding, and cohort discovery in electronic health records (EHRs) \cite{bai2018ehr, glicksberg2018automated}; they also surface relevant evidence in retrieval-augmented generation (RAG) systems \cite{lewis2020retrieval}; and boost recall as well as ranking in biomedical information retrieval \cite{yeganova2020better}. By mitigating lexical variability and contextual ambiguity, embeddings support efficient similarity search, patient clustering, and reasoning across diverse medical corpora. Consequently, embedding techniques have become integral in EHR analytics, unifying structured and unstructured data to improve patient stratification and outcome prediction \cite{zhang2020combining}.

Pre-trained transformer embeddings are now central to NLP. BERT \cite{devlin2018bert} introduced bidirectional masked language modeling and next-sentence prediction, enabling deep contextual understanding. SciBERT \cite{beltagy2019scibert} was built on BERT by pretraining on a scientific-article corpus, boosting performance in biology and chemistry tasks. While BERT and SciBERT excel at token-level representations, Sentence-BERT (SBERT) \cite{reimers2019sentence} uses Siamese and triplet networks to produce semantically optimized sentence embeddings. The E5 family \cite{wang2022text} further advances this by applying large-scale, weakly supervised contrastive learning over billions of text-pair examples. Emerging instruction tuned and multilingual models continue this trajectory, extending transformer embeddings to broader tasks and languages \cite{Chen2024M3Embed}.

Recent work merges unsupervised and supervised contrastive objectives. GTE \cite{li2023towards} uses large‐scale unsupervised contrastive pretraining followed by supervised fine‐tuning. Nomic-Embed \cite{nussbaum2024nomic} combines masked language modeling and contrastive learning over contexts up to 8,192 tokens with FlashAttention, gradient caching, and DeepSpeed. These steps shift from bidirectional pretraining toward specialized contrastive methods, motivating our medical embedding benchmark.

In the medical domain, transformer embeddings have been tailored to clinical data: Med-BERT \cite{rasmy2021med} adds cross-visit pretraining on EHRs from Cerner and Truven MarketScan; ClinicalBERT \cite{huang2019clinicalbert} continues pretraining on MIMIC-III notes; BioBERT \cite{lee2020biobert} augments pretraining with PubMed abstracts and PMC full texts; ExBERT \cite{tai2020exbert} integrates a module for out-of-vocabulary medical terms. These domain-specialized methods further inform our medical embedding benchmark. Notably, the recent GatorTron model scaled a clinical transformer to 8.9B parameters on 90B words of de-identified notes and biomedical text, achieving state-of-the-art results on clinical NLI, STS, and QA tasks \cite{Yang2022GatorTron}. This demonstrates that increasing model size and in-domain pretraining data can substantially boost performance in clinical NLP.

Complementing these domain-specialized strategies, contrastive and self-supervised methods have refined biomedical embeddings: BioSimCSE \cite{kanakarajan2022biosimcse} fine tunes sentences with contrastive loss; Abro et al. \cite{abro2024self} apply self-supervised contrastive tuning to ClinicalBioBERT; Min et al. \cite{min2024adaption} integrate ChatGPT-generated paraphrases into a contrastive loop; NoteContrast \cite{kailas2023notecontrast} aligns clinical notes with ICD-10 codes; MICOL \cite{zhang2022metadata} uses metadata for hierarchical contrastive learning; MedEmbed \cite{balachandran2024medembed} trains on PubMed triplets with hard negatives. Building on these, BioLORD-2023 couples UMLS knowledge graph augmentation with contrastive learning and self-distillation, setting new state-of-the-art on biomedical sentence similarity and zero-shot concept mapping tasks \cite{Remy2023BioLORD}.

Despite rapid advances in general text embeddings, evaluations of medical‐domain models remain fragmented and narrow. Most benchmarks cover only a few clinical datasets or task types, leaving critical gaps in our understanding of model behavior across real‐world healthcare scenarios. The Massive Text Embedding Benchmark (MTEB) \cite{muennighoff2022mteb} addresses this by unifying 58 datasets in 112 languages over eight core tasks, including bitext mining, classification, clustering, pair classification, reranking, retrieval, semantic textual similarity, and summarization via an open-source design and Hugging Face leaderboard. However, MTEB’s medical subsection comprises only 12 unevenly sized datasets (eight in English), limiting its utility for rigorous medical embedding assessment. The Chemical Text Embedding Benchmark (ChemTEB) \cite{kasmaee2024chemteb} demonstrates how specialized suites can capture domain‐specific challenges through tasks like chemical document classification, retrieval, and bitext mining, inspiring our approach to medical text embedding evaluation.

Subsequent efforts have targeted scalable and retrieval‐focused evaluation in clinical contexts. Soffer et al. \cite{soffer2024scalable} tested 30 top MTEB models on heterogeneous corpora, PubMed abstracts, LLaMA‐3‐70B–generated synthetic EHRs, Mount Sinai patient records, and MIMIC-IV notes, distinguishing “Short Tasks” (e.g., triage notes) from “Long Tasks” (e.g., full patient histories). While revealing length‐dependent performance trends, this framework remains closed‐source and limited to matching tasks. HEmTEB (Health Embedding Benchmark) \cite{hemteb2024} refines retrieval evaluation by curating clinician‐supervised datasets across ten specialties and assessing cross‐lingual capabilities, yet it too omits classification, clustering, and semantic similarity assessments. Parallel evaluations on specialized tasks (e.g., biomedical terminology mapping) similarly find that no single model excels uniformly across all semantic alignment challenges, underscoring the need for broader benchmarks \cite{Lahiri2024TerminologyBenchmark}.

These specialized frameworks underscore the value of targeted benchmarks but also highlight the need for a more expansive, standardized medical embedding evaluation. We therefore propose the Medical Text Embedding Benchmark, a large‐scale, open‐source suite spanning classification, clustering, retrieval, semantic similarity, and beyond, drawn from diverse clinical and biomedical corpora, to provide a comprehensive, transparent, and actionable performance assessment for medical NLP models.

\section{Methodology}

\subsection{Sources of data for MedTEB and MedTE}

Our corpus integrates complementary strands of biomedical knowledge from peer-reviewed literature, preprint archives, real-world clinical narratives, structured registries, human-curated encyclopedic resources, standardized question–answer benchmarks, and specialized training corpora. We begin with {PubMed} \cite{pubmed-site}, a comprehensive catalogue of biomedical abstracts, and its full-text counterpart {PubMed Central (PMC)} \cite{pmc-site}, anchoring our collection in rigorously vetted scientific discourse. To capture authentic clinical language we incorporate {MIMIC-IV} \cite{johnson2023mimic}, a de-identified EHR dataset rich in clinical notes, discharge summaries, and structured patient data reflective of bedside documentation. Prospective study descriptions from {ClinicalTrials.gov} \cite{ctgov-site} contribute structured trial metadata, objectives, interventions, and eligibility criteria, extending our coverage to both ongoing and completed research across therapeutic domains. For rapid insight into emerging findings we harvest life-science preprints from {bioRxiv} \cite{biorxiv-2019} and clinically oriented preprints from {medRxiv} \cite{medrxiv-site}, whose detailed metadata support fine-grained filtering and trend analysis.

Human-curated {Wikipedia} medical pages provide systematically structured descriptions of diseases, diagnostics, and therapies, bridging professional terminology and lay explanations. To evaluate reasoning and factual recall we include multiple-choice question sets: {MedMCQA}, derived from Indian medical entrance examinations \cite{pal2022medmcqa}, and {MedQA}, reflecting US medical licensing assessments \cite{jin2021disease}. Finally, the {MedQuAD} corpus supplies 47457 question–answer pairs sourced from authoritative outlets such as the National Cancer Institute, enriching the benchmark with diverse consumer-health inquiries \cite{BenAbacha-BMC-2019}. Additionally, we employ {Trec-covid} \cite{voorhees2021trec}, {NFCorpus} \cite{boteva2016}, and {CUREv1} \cite{hemteb2024} solely for model training. Together, these resources yield a unified, high-coverage dataset that interleaves structured registries, scholarly publications, clinical text, and educational materials to form a robust foundation for training and evaluating medical language models across retrieval, comprehension, and reasoning tasks.

For both model training and benchmarking, we used these data sources. However, to ensure unbiased evaluation, all benchmark data were removed from the training data. This separation guarantees that the model’s performance on downstream tasks is not influenced by any overlap between the training and benchmark datasets.

\subsection{MedTE Training}

\medskip
\noindent\textbf{Data Preprocessing}

For every corpus, we derive semantically aligned \emph{positive pairs}, two text fragments that convey the same underlying meaning.  The pairing heuristics for each data source are listed in Table~\ref{tab:data_preprocess_adjusted_updated}.  All texts are first lower‑cased and then tokenised with the GTE tokenizer (a \texttt{bert‑base‑uncased} variant) to ensure a uniform sub‑word vocabulary.

\begin{table*}[htbp]
\centering

\begin{tabular}{lp{0.55\linewidth}r}
\toprule
\textbf{Data Source} & \textbf{Sentence Pair Definition} & \textbf{Number of samples} \\
\midrule
PubMed               & Article \textit{title} $\leftrightarrow$ {abstract} sentence \#1                   & 572\,300 \\
bioRxiv / medRxiv    & Pre-print {title} $\leftrightarrow$ {abstract} sentence \#1                   & 231\,400 \\
MIMIC-IV             & {History of Present Illness} $\leftrightarrow$ {Chief Complaint}            & 311\,400 \\
ClinicalTrials.gov   & Study {title} $\leftrightarrow$ {detailed description}                      & 378\,800 \\
MedMCQA              & Exam {question} $\leftrightarrow$ {answer explanation}                      & 151\,000 \\
MedQA                & Exam {question} $\leftrightarrow$ {answer explanation}                      &   5\,300 \\
MedQuAD              & User {question} $\leftrightarrow$ authoritative {answer passage}            &   8\,500 \\
TREC-COVID           & Search {query} $\leftrightarrow$ relevant {passage}                         & 129\,200 \\
NF-corpus            & Information-need {query} $\leftrightarrow$ relevant {document snippet}      &   3\,600 \\
CURE-V1              & Clinical {query} $\leftrightarrow$ supporting {evidence sentence}           & 242\,300 \\
\midrule
\textbf{Total}       & & 2\,033\,800 \\
\bottomrule
\end{tabular}
\caption{Positive-pair construction for MedTE's unsupervised contrastive learning (numbers rounded).}
\label{tab:data_preprocess_adjusted_updated}
\end{table*}

\medskip
\noindent\textbf{Training Objectives and Model Architecture}

We adopt a single-stage, \textit{unsupervised} contrastive learning strategy built on the GTE-base architecture.  The transformer backbone remains identical to the public GTE implementation, but all parameters are updated during training to specialise the encoder for biomedical and clinical language. Given a mini--batch $\mathcal{B}={(x_i, x_i^+)}{i=1}^{N}$ of $N$ positive sentence pairs, we maximise the InfoNCE objective
\begin{equation}
\mathcal{L}{\text{InfoNCE}};=; - \frac{1}{N}\sum_{i=1}^N \log \frac{\exp\big(\operatorname{sim}(z_i, z_i^+)/\tau\big)}{\sum\limits_{j=1}^{N}\exp\big(\operatorname{sim}(z_i, z_j^+)/\tau\big)},\label{eq:infonce}
\end{equation}
where $z=\operatorname{MeanPool}(\text{GTE}(x))$ is the sentence embedding, $\operatorname{sim}(\cdot,\cdot)$ denotes cosine similarity, and $\tau$ is a learned temperature parameter.  No manually labelled negatives are required: all non‑matching sentences in the same batch act as implicit negatives, enabling scalable self‑supervision. To keep domain shift under control, the data loader assembles batches from a single corpus at a time, so every negative originates from the same data source as its corresponding positive pair.

We retain GTE's mean pooling over the final hidden states.  Early experiments with max pooling and CLS pooling showed no measurable gains and were hence discarded. The developers of GTE did not incorporate an MLM phase, training the model exclusively with contrastive learning. Furthermore, in our own preliminary experiments, adding an MLM stage before contrastive learning significantly added to training time without yielding any downstream performance gains. Consequently, we omitted the MLM step and trained the model end-to-end using only the contrastive objective.

\medskip
\noindent\textbf{Training Setup}

Training is performed on {2,033,800} positive pairs (\ref{tab:data_preprocess_adjusted_updated}) drawn proportionally from all corpora. Each optimization step processes a batch of {1024} pairs, and we train for {8000} steps, which we found sufficient for convergence. We use the {AdamW} \cite{loshchilov2017decoupled} optimizer with weight decay of $0.01$; the learning rate is linearly warmed up to its peak over the first 1,000 steps and then decays according to a cosine schedule. Mixed precision ({bfloat16}) is enabled to reduce memory footprint, and gradient checkpointing is activated to fit larger batches without compromising sequence length. We also leverage {DeepSpeed}\cite{rasley2020deepspeed} for efficient large‐batch training and accelerated throughput. At inference time, embeddings are $L_2$-normalized so that inner-product search is equivalent to cosine similarity.

\subsection{MedTEB Development}

In our benchmark, we assembled 51 datasets across four categories, Classification, Clustering, Pair Classification, and Retrieval, to comprehensively evaluate the representational power of embedding models. For Classification, given a dataset \(\mathcal{D}=\{(x_i,y_i)\}_{i=1}^N\), we fine-tune the embedding model on the training split \(\mathcal{D}_{\mathrm{train}}\), feed the resulting sentence embeddings into a logistic regression classifier trained on the same split, and report macro-averaged \(F_1\) on the test split \(\mathcal{D}_{\mathrm{test}}\). In Clustering tasks, each input \(x_i\) is embedded into \(z_i\in\mathbb{R}^d\) and Mini-Batch \(k\)-means (batch size 32) is applied to produce cluster labels \(\hat c_i\), with clustering quality measured against true labels \(c_i\) using the V-measure. For Pair Classification, given \(\mathcal{D}=\{(x_i^{(1)},x_i^{(2)},y_i)\}_{i=1}^N\) with binary labels \(y_i\), we compute four similarity or distance scores (cosine, Euclidean, Manhattan, and dot product) per pair, select a threshold \(\tau\) on the training data to maximize \(F_1\), and define predictions \(\hat y_i\) accordingly, reporting the best \(F_1\) across all metrics. Finally, Retrieval tasks consist of query sets \(\mathcal{Q}\), document corpora \(\mathcal{D}\), and relevance pairs \(\mathcal{R}\subseteq\mathcal{Q}\times\mathcal{D}\); we embed every query and document into vectors \(z_{q_j}\) and \(z_{d_k}\), rank documents by descending cosine similarity \(\cos(z_{q_j},z_{d_k})\), and measure performance with nDCG@10 averaged over queries. This design enables a holistic analysis of embedding quality across supervised and unsupervised scenarios, capturing semantic grouping, pairwise similarity discrimination, and practical retrieval capabilities in information-seeking tasks. Table~\ref{tab:checkmark_table_extended} provides an overview of all datasets and their properties.

\begin{table*}[ht]
  \centering
  
  \begin{adjustbox}{max width=\textwidth}
  \begin{tabular}{lcccccccccc}
    \toprule
    \textbf{Task}       & \textbf{MIMIC-IV} & \textbf{PMC} & 
\textbf{PubMed} & \textbf{Wikipedia} & \textbf{MedQA} & \textbf{MedMCQA} & \textbf{MedQUAD} & \textbf{medRxiv} & \textbf{bioRxiv} & \textbf{Total} \\
    \midrule
    Classification      & \cmark & \cmark & \cmark & \cmark & \xmark & \xmark & \xmark & \xmark & \xmark &  15 \\
    Clustering          & \cmark & \cmark & \xmark & \cmark & \xmark & \xmark & \xmark & \xmark & \xmark & 12 \\
    Pair Classification & \cmark & \xmark & \xmark & \xmark & \cmark & \cmark & \xmark & \cmark & \xmark & 12 \\
    Retrieval           & \cmark & \xmark & \cmark & \cmark & \cmark & \cmark & \cmark & \cmark & \cmark & 12 \\
    \bottomrule
  \end{tabular}
  \end{adjustbox}
  \caption{Presence of Data Sources Across MedTEB Tasks}
  \label{tab:checkmark_table_extended}
\end{table*}

We curated and preprocessed diverse text corpora with fixed train/validation/test splits and formatted each dataset for classification, clustering, pair classification, or retrieval; below, we outline the specific curation methods for each task type.

\medskip
\noindent\textbf{Classification \& Clustering.} 

For each dataset, we extracted the introductory or abstract portion of every document as the input text.  Labels were then assigned by one of three methods: (i) the author’s own keywords, (ii) the document’s parent category in its source hierarchy, or (iii) the search term used to retrieve the document when no explicit label was available.  This process yields a clean set of \(\langle\text{text},\text{label}\rangle\) pairs that serve both classification and clustering experiments.

\medskip
\noindent\textbf{Pair Classification.}  

In addition to a few native question–answer corpora, we generated positive pairs automatically.  Given a sentence \(s\), we prompted GPT-4o to produce a semantically equivalent (or highly related) paraphrase \(s^{+}\).  To create challenging negatives, we computed BERT embeddings for all candidate sentences, identified the 64 nearest neighbours of \(s^{+}\) (excluding itself), and randomly sampled one as the negative example \(s^{-}\).  Early tests with purely random negatives resulted in near-perfect F\(_1\) scores for many models; our hard-negative mining restores the discriminative difficulty needed for robust evaluation.

\medskip
\noindent\textbf{Retrieval.}  

We reused the same corpus by treating each original sentence as a “document” and its GPT-4o paraphrase (or, in QA datasets, the question) as a query.  This guarantees that every \(\langle\text{query},\text{document}\rangle\) pair is relevant, while the rest of the corpus acts as distractors.  By mirroring the Pair Classification setup, we ensure consistent difficulty across both tasks.

\medskip
We evaluated a diverse collection of embedding models that differ along two key dimensions, domain (general versus medical) and training objective (contrastive learning versus standard pretraining). Table~\ref{tab:model_specs} summarizes each model’s size, parameter count, maximum context length, embedding dimensionality, and whether it was trained with a contrastive objective. The non-medical group includes popular transformers such as BGE, SciBERT, and MiniLM variants, while the medical group comprises specialized models like ClinicalBERT, BioSimCSE, and our own MedTE.

\begin{table*}[ht]
\centering

\begin{adjustbox}{max width=\textwidth}
\begin{tabular}{lccccc}
\toprule
\textbf{Model Name} & \textbf{Size (MB)} & \textbf{Params (M)} & \textbf{Context} & \textbf{Emb.\ Dim} & \textbf{Cons. Lear.} \\
\midrule
\multicolumn{6}{l}{\textit{Non‑medical models}}\\
\midrule
BAAI BGE Base En V1.5                           & 418 & 109.48 & 512 & 768 & Yes \\
AllenAI SciBERT Scivocab Uncased                & 422 & 109.92 & 512 & 768 & No  \\
Google BERT Base Uncased                        & 420 & 109.48 & 512 & 768 & No  \\
Intfloat E5 Base                                & 418 & 109.48 & 512 & 768 & Yes \\
Nomic AI Nomic Embed Text V1                    & 522 & 136.73 & 819 & 768 & Yes \\
Nomic AI Nomic Embed Text V1 Unsupervised       & 522 & 136.73 & 512 & 768 & Yes \\
Sentence‑Tfrs All‑MiniLM‑L6‑v2                  &  87 &  22.71 & 512 & 384 & Yes \\
Sentence‑Tfrs All‑MPNet‑Base‑v2                 & 418 & 109.49 & 512 & 768 & Yes \\
Thenlper GTE Base                               & 209 & 109.48 & 512 & 768 & Yes \\
\midrule
\multicolumn{6}{l}{\textit{Medical models}}\\
\midrule
BioNLP BlueBERT PubMed MIMIC Uncased L‑12 H‑768 A‑12 & 420 & 109.48 & 512 & 768 & No  \\
Abhinand MedEmbed Base                           & 420 & 109.48 & 512 & 768 & Yes \\
Emily Alsentzer Bio‑ClinicalBERT                & 416 & 108.31 & 512 & 768 & No  \\
Kamalkraj BioSimCSE BioLinkBERT Base            & 413 & 108.23 & 512 & 768 & Yes \\
Malteos SciNCL                                  & 419 & 109.92 & 512 & 768 & Yes \\
MedicalAI ClinicalBERT                          & 517 & 134.73 & 512 & 768 & No  \\
Microsoft BiomedBERT Base Uncased Abs+Fulltext  & 420 & 109.48 & 512 & 768 & No  \\
\midrule
 MedTE Cl15 Step 8000                 & 438 & 109.48 & 512 & 768 & Yes \\
\bottomrule
\end{tabular}
\end{adjustbox}
\caption{Specifications for Evaluated Embedding Models on MedTEB, with Domain and Contrastive‑Learning Indicators}
\label{tab:model_specs}
\end{table*}

\section{Results and Discussion}

\begin{table*}[htbp]
  \footnotesize                       
  \setlength{\tabcolsep}{3pt}         
  \begin{adjustbox}{max width=\textwidth}
    \centering

    \small
    \begin{tabular}{lccccccc}
      \toprule
      \textbf{Model} & \textbf{Classification} & \textbf{Clustering} & \textbf{Pair Cls.} & \textbf{Retrieval} & \textbf{AvgType} & \textbf{AvgAll} & \textbf{EvalTime} \\
      \midrule
      \multicolumn{8}{l}{\textit{Non-medical models}}\\
      \midrule
      BAAI Bge Base En V1.5 & 0.69 ± 0.22 & 0.28 ± 0.28 & 0.67 ± 0.15 & 0.38 ± 0.36 & 0.505 & 0.516 & 102.78 \\
      AllenAI Scibert Scivocab Uncased & 0.60 ± 0.20 & 0.06 ± 0.05 & 0.64 ± 0.12 & 0.06 ± 0.07 & 0.341 & 0.356 & 99.26 \\
      Google BERT Base Uncased & 0.58 ± 0.20 & 0.10 ± 0.11 & 0.63 ± 0.11 & 0.08 ± 0.10 & 0.348 & 0.362 & 102.08 \\
      Intfloat E5 Base & 0.68 ± 0.22 & 0.24 ± 0.25 & 0.67 ± 0.15 & 0.33 ± 0.33 & 0.479 & 0.490 & 102.51 \\
      Nomic AI Nomic Embed Text V1 & 0.69 ± 0.22 & 0.28 ± 0.28 & 0.67 ± 0.15 & 0.38 ± 0.36 & 0.503 & 0.511 & 112.79 \\
      Nomic AI Nomic Embed Text V1 Unsupervised & 0.69 ± 0.22 & 0.29 ± 0.21 & 0.68 ± 0.16 & 0.40 ± 0.35 & 0.515 & 0.525 & 90.15 \\
      Sentence-Transformers All MiniLM L6 V2 & 0.68 ± 0.21 & 0.28 ± 0.17 & 0.66 ± 0.14 & 0.34 ± 0.33 & 0.489 & 0.501 & \textbf{20.87} \\
      Sentence-Transformers All MPNet Base V2 & 0.70 ± 0.22 & 0.30 ± 0.21 & 0.67 ± 0.15 & 0.35 ± 0.34 & 0.502 & 0.514 & 54.98 \\
      Thenlper GTE Base & 0.70 ± 0.22 & 0.28 ± 0.24 & 0.69 ± 0.16 & 0.41 ± 0.35 & 0.518 & 0.529 & 94.98 \\
      \midrule
      \multicolumn{8}{l}{\textit{Medical models}}\\
      \midrule
      Abhinand MedEmbed Base & 0.69 ± 0.22 & 0.36 ± 0.28 & 0.68 ± 0.16 & 0.39 ± 0.35 & 0.529 & 0.539 & 53.60 \\
      BioNLP Bluebert PubMed Mimic Uncased L-12 H-768 A-12 & 0.62 ± 0.21 & 0.09 ± 0.11 & 0.63 ± 0.11 & 0.07 ± 0.07 & 0.351 & 0.367 & 52.23 \\
      EmilyAlsentzer Bio ClinicalBERT & 0.59 ± 0.21 & 0.06 ± 0.06 & 0.63 ± 0.11 & 0.05 ± 0.05 & 0.334 & 0.349 & 104.01 \\
      Kamalkraj BioSimCSE BioLinkBERT Base & 0.64 ± 0.22 & 0.23 ± 0.24 & 0.66 ± 0.14 & 0.27 ± 0.31 & 0.449 & 0.460 & 53.83 \\
      Malteos SciNCL & 0.69 ± 0.22 & 0.34 ± 0.26 & 0.66 ± 0.14 & 0.30 ± 0.30 & 0.498 & 0.509 & 98.57 \\
      MedicalAI ClinicalBERT & 0.60 ± 0.21 & 0.10 ± 0.10 & 0.63 ± 0.11 & 0.06 ± 0.06 & 0.346 & 0.366 & 43.07 \\
      Microsoft BiomedNLP BiomedBERT Base Uncased Abstract Fulltext & 0.61 ± 0.20 & 0.12 ± 0.17 & 0.64 ± 0.12 & 0.13 ± 0.17 & 0.374 & 0.388 & 49.34 \\
      \midrule
      MedTE Cl15 Step 8000 & \textbf{0.72 ± 0.23} & \textbf{0.38 ± 0.24} & \textbf{0.74 ± 0.17} & \textbf{0.45 ± 0.32} & \textbf{0.569} & \textbf{0.578} & 54.63 \\
      \bottomrule
    \end{tabular}
  \end{adjustbox}
  \caption{Performance of embedding models on various MedTEB tasks (mean ± standard deviation)}
  \label{tab:embedding_performance}
\end{table*}

Table~\ref{tab:embedding_performance} gives an arranged view of the empirical benchmarks spanning \emph{classification}, \emph{clustering}, \emph{pair classification}, and \emph{retrieval}. Our self-supervised model emerges as the top performer in \emph{every} task family, validating the effectiveness of a pure contrastive regimen in highly specialised clinical language.


\subsection{Overall performance}

Averaged across all 16 task metrics, Our model attains a mean score of \(0.578\), eclipsing the nearest competitor at \(0.539\) (Abhinand MedEmbed Base) and establishing a new state of the art for medical sentence representation. These results underscore that in specialized domains such as healthcare, self-supervised contrastive learning is not optional but decisive, future medical NLP systems should therefore prioritize contrastive frameworks to achieve consistent improvements in classification, clustering, and retrieval.  

\subsection{Classification}
Our model achieves an F\(_1\) score of \(0.72 \pm 0.23\), outperforming both the strongest non-medical baseline (MPNet Base V2 and Thenlper GTE Base , \(0.70\pm0.22\)) and the leading clinical encoder (Abhinand MedEmbed Base and SciNCL, \(0.69\pm0.22\)). In practical terms, this \(+0.02\)–\(+0.03\) margin translates to hundreds fewer misclassifications in downstream triage workflows, since each point of F\(_1\) corresponds to hundreds of patient records.

\subsection{Clustering}
Using the V-measure, our model records \(0.38\pm0.24\), a gain of \(+0.08\) over the top non-medical contrastive model (MPNet Base V2, \(0.30\pm0.21\)) and \(+0.02\) above the best domain-tuned rival (Abhinand MedEmbed Base , \(0.36\pm0.28\)).

\subsection{Pair Classification and Retrieval}

For semantically equivalent sentence detection,  posts an F\(_1\) of \(0.74\pm0.17\), Thenlper GTE Base by \(+0.05\) and MedEmbed Base by \(+0.06\). In retrieval, it achieves nDCG@10 \(=0.45\pm0.32\), improving over Thenlper GTE Base by \(+0.04\) and the strongest clinical baseline (Abhinand MedEmbed Base ) by \(+0.06\). These gains reflect a denser, better-calibrated embedding space that surfaces relevant clinical evidence with fewer query reformulations.


\subsection{Effect of contrastive learning}

General-purpose BERT-style encoders (e.g., Google BERT Base, SciBERT) consistently lag on our medical benchmarks (e.g., In classification the best BERT Base models achieves: F$_1 = 0.60 \pm 0.20$ and in  clustering V-measure = $0.10 \pm 0.11$) because they lack a contrastive fine tuning stage. Off the shelf clinical models such as BioClinicalBERT, ClinicalBERT, BlueBERT, and BiomedBERT, despite large scale biomedical pretraining, also underperform (The best BERT Base model In clustering achieves V-measure $\approx 0.12\text{--}0.17$ and in retrieval nDCG@10 $\approx 0.13\text{--}0.17$), showing that domain-specific pretraining alone does not guarantee fine-grained semantic discrimination. In particular, retrofitting a general transformer with contrastive learning (e.g. SciNCL) can surpass many purpose built clinical embeddings, and exclusive GTE style contrastive training further refines the embedding manifold to respect subtle concept hierarchies (e.g. 'angina' $\rightarrow$ 'ischaemic heart disease') while remaining robust across note types.

\section{Analysis}

\subsection{Per Source Performance on MedTE}
As explained in the data section, we created our benchmark using a diverse set of medical text data from different sources. In table \ref{tab:embedding_performance_src}, we visualized the average performance of models on different sources of data. Both Abhinand MedEmbed Base and Thenlper GTE Base achieve particularly strong performance on PubMed (0.89–0.90) and on BioRxiv (0.89), substantially outpacing models such as SciBERT (PubMed 0.50; BioRxiv 0.18) and BioClinicalBERT (PubMed 0.47; BioRxiv 0.10). Ours further refines this trend, achieving the highest MIMIC‐IV score (0.61) and maintaining robust performance across other clinical sources (Clinical Trials 0.81; MedRxiv 0.74). In comparison, general embeddings such as All MiniLM L6 V2 and MPNet Base V2 hover around 0.51–0.55 on MIMIC‐IV and just 0.64–0.67 on Clinical Trials, underscoring the importance of specialized pretraining for domain‐specific tasks.

The MedMCQA and MedQA tasks involve multi‐option question answering, which poses a significant challenge for embedding models to capture the relationship between a question and its correct answer. All models converge to a 0.39 average on these datasets, highlighting this difficulty. Nevertheless, Ours outperforms competitors on MedQA, likely attributable to its extensive pretraining on a diverse set of medical data. On the general‐domain Wikipedia task, domain‐adapted models incur only a minor drop, Ours achieves 0.45 and Thenlper GTE Base reaches 0.43, outperforming many general embeddings (All MiniLM L6 V2 at 0.41; BERT Base at 0.26). Overall, these results confirm that our model outperforms existing embeddings across specialized biomedical evaluations while preserving, and in some cases enhancing, performance on broader‐scope information sources.

\begin{figure}[htpb]
  \centering
  \includegraphics[width=\columnwidth]{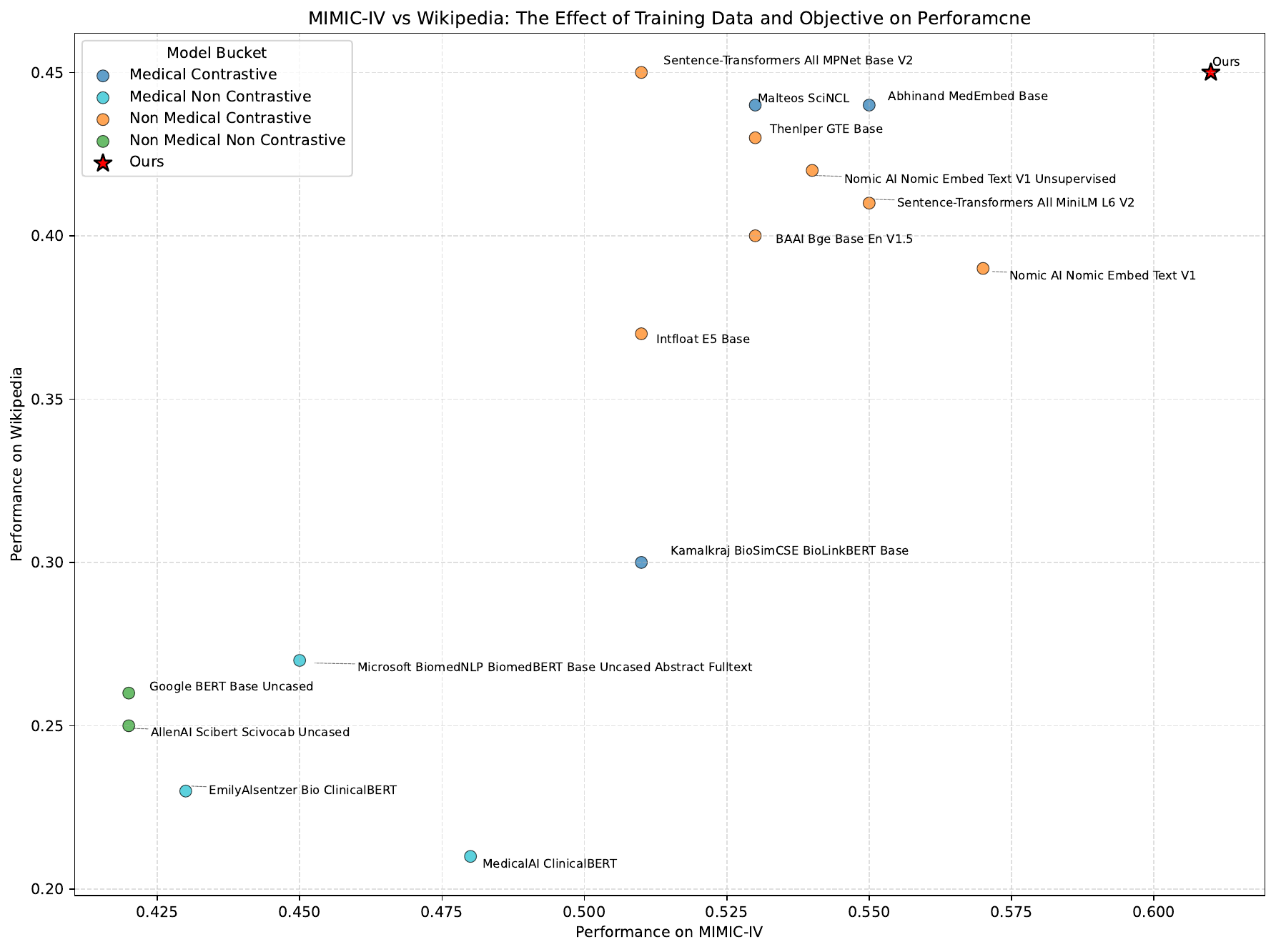}
  \caption{Comparison of model performance on Wikipedia versus MIMIC‑IV}
  \label{fig:wiki_vs_mimc}
\end{figure}

Figure~\ref{fig:wiki_vs_mimc} illustrates the stark contrast in model performance between MIMIC-IV and Wikipedia tasks. These two datasets differ fundamentally in content and style: MIMIC-IV consists of specialized, technical clinical records, whereas Wikipedia entries are broadly educational and follow a more structured, encyclopedic format. Moreover, embedding models are often pretrained on Wikipedia data, giving them a natural advantage on that domain. In contrast, the professional and jargon-heavy language of MIMIC-IV poses a greater challenge, which is reflected in the relative performance gap. A very interesting observation is that Nomic Embed Text V1, while being a general model, is better than any other models except for ours, which shows their model is even better than medical models. We can also see the performance of models clustered into two groups based on whether they have contrastive learning or not. 
We additionally compare the Wikipedia and MIMIC-IV datasets against PubMed in the Appendix.

\begin{table*}[htbp]
  \footnotesize                       
  \setlength{\tabcolsep}{3pt}         
  \begin{adjustbox}{max width=\textwidth}
    \centering
    \small

    \begin{tabular}{lcccccccccc}
      \toprule
      \textbf{Model} & BioRxiv & Clinical Trials & MIMIC-IV & MedMCQA & MedQA & MedQuAD & MedRxiv & PMC & PubMed & Wikipedia \\
      \midrule
      \multicolumn{11}{l}{\textit{Non-medical models}}\\
      \midrule
      BAAI Bge Base En V1.5                           & 0.89 & 0.68 & 0.53 & \textbf{0.39} & 0.39 & 0.40 & 0.78 & \textbf{0.28} & 0.88 & 0.40 \\
      AllenAI Scibert Scivocab Uncased                & 0.18 & 0.37 & 0.42 & \textbf{0.39} & 0.39 & 0.05 & 0.06 & 0.27 & 0.50 & 0.25 \\
      Google BERT Base Uncased                        & 0.27 & 0.37 & 0.42 & \textbf{0.39} & 0.39 & 0.06 & 0.13 & 0.26 & 0.51 & 0.26 \\
      Intfloat E5 Base                                & 0.85 & 0.65 & 0.51 & \textbf{0.39} & 0.39 & 0.38 & 0.66 & 0.27 & 0.79 & 0.37 \\
      Nomic AI Nomic Embed Text V1                    & \textbf{0.90} & 0.67 & 0.57 & \textbf{0.39} & 0.39 & 0.37 & \textbf{0.82} & \textbf{0.28} & 0.88 & 0.39 \\
      Nomic AI Nomic Embed Text V1 Unsupervised       & \textbf{0.90} & 0.71 & 0.54 & \textbf{0.39} & 0.39 & 0.37 & \textbf{0.82} & 0.27 & 0.88 & 0.42 \\
      Sentence-Transformers All MiniLM L6 V2          & 0.85 & 0.64 & 0.55 & \textbf{0.39} & 0.39 & 0.35 & 0.68 & \textbf{0.28} & 0.82 & 0.41 \\
      Sentence-Transformers All MPNet Base V2         & 0.85 & 0.67 & 0.51 & \textbf{0.39} & 0.39 & 0.39 & 0.69 & \textbf{0.28} & 0.86 & \textbf{0.45} \\
      Thenlper GTE Base                               & 0.89 & 0.72 & 0.53 & \textbf{0.39} & 0.39 & \textbf{0.42} & 0.78 & \textbf{0.28} & \textbf{0.90} & 0.43 \\
      \midrule
      \multicolumn{11}{l}{\textit{Medical models}}\\
      \midrule
      Abhinand MedEmbed Base                          & 0.89 & 0.69 & 0.55 & \textbf{0.39} & 0.39 & \textbf{0.42} & 0.77 & \textbf{0.28} & 0.89 & 0.44 \\
      BioNLP BlueBERT PubMed MIMIC Uncased L-12 768 12 & 0.09 & 0.39 & 0.45 & \textbf{0.39} & 0.39 & 0.10 & 0.05 & 0.26 & 0.48 & 0.25 \\
      EmilyAlsentzer Bio ClinicalBERT                 & 0.10 & 0.37 & 0.43 & \textbf{0.39} & 0.39 & 0.06 & 0.06 & 0.26 & 0.47 & 0.23 \\
      Kamalkraj BioSimCSE BioLinkBERT Base            & 0.77 & 0.56 & 0.51 & \textbf{0.39} & 0.39 & 0.20 & 0.60 & 0.26 & 0.82 & 0.30 \\
      Malteos SciNCL                                  & 0.78 & 0.63 & 0.53 & \textbf{0.39} & 0.39 & 0.25 & 0.55 & 0.27 & 0.77 & 0.44 \\
      MedicalAI ClinicalBERT                          & 0.13 & 0.38 & 0.48 & \textbf{0.39} & 0.39 & 0.07 & 0.07 & 0.26 & 0.45 & 0.21 \\
      Microsoft BiomedBERT (abstract+full-text)       & 0.48 & 0.40 & 0.45 & \textbf{0.39} & 0.39 & 0.07 & 0.24 & 0.27 & 0.61 & 0.27 \\
      \midrule
      MedTE Cl15 Step 8000                         & 0.86 & \textbf{0.81} & \textbf{0.61} & \textbf{0.39} & \textbf{0.45} & \textbf{0.42} & 0.74 & 0.27 & 0.87 & \textbf{0.45} \\
      \bottomrule
    \end{tabular}
  \end{adjustbox}
  \caption{Performance of Embedding Models on Various Data Sources}
  \label{tab:embedding_performance_src}
\end{table*}

\subsection{Disease clustering}
\begin{figure}[htpb]
  \centering
  \includegraphics[width=\columnwidth]{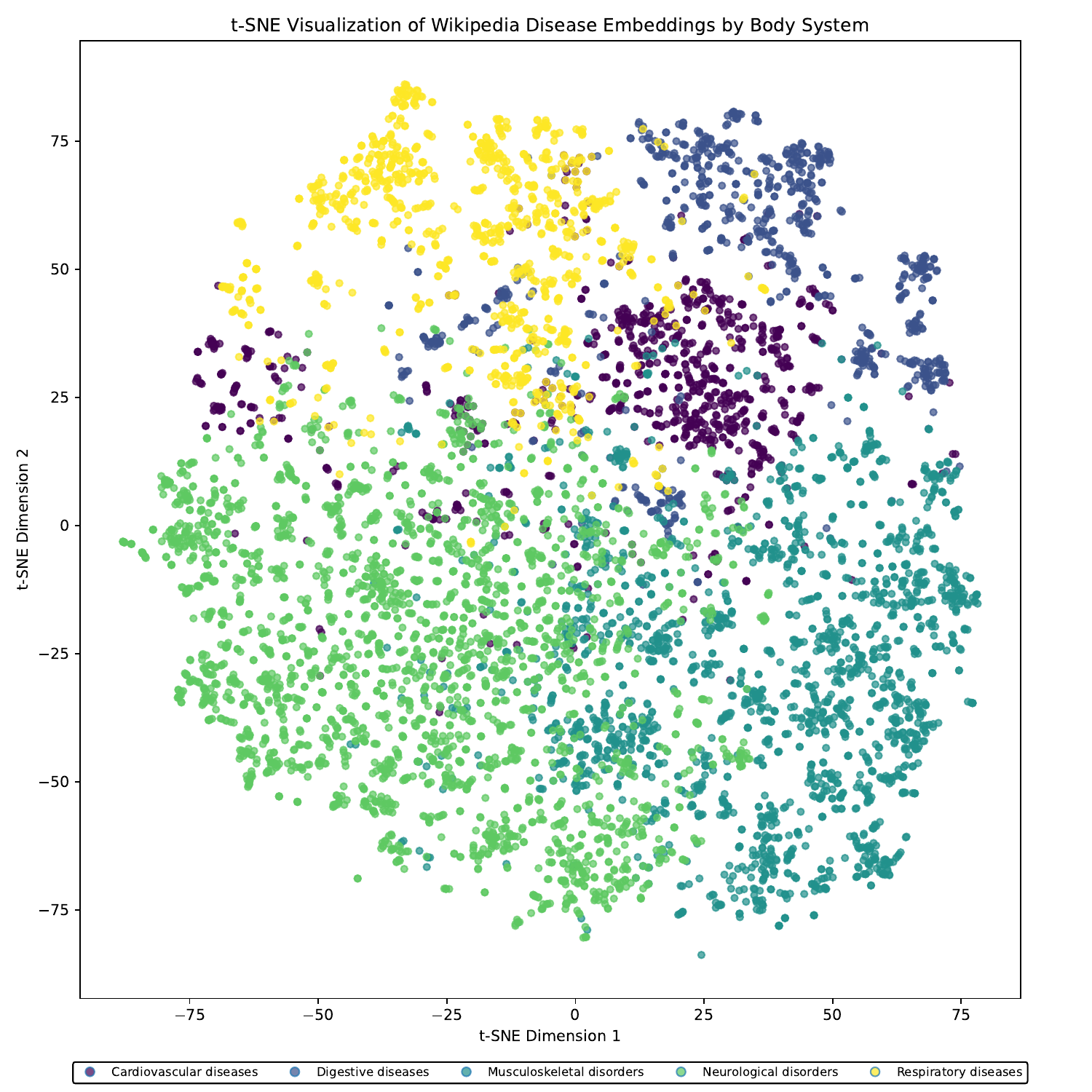}
  \caption{t‑SNE Visualization of Wikipedia Disease Embeddings by Body System.}
  \label{fig:tsne_wiki_disease_by_system_filtered}
\end{figure}

As previously demonstrated, our model outperforms all baselines on the clustering task (Table \ref{tab:embedding_performance}), confirming its unrivaled ability to group semantically related medical texts. Moreover, the qualitative t-SNE visualization in Figure \ref{fig:tsne_wiki_disease_by_system_filtered}, which projects Wikipedia-derived embeddings into two dimensions and color-codes each point by disease and body system, further corroborates this finding. Here, each disease category forms a distinct, tightly knit cluster with minimal overlap, illustrating the model’s remarkable sensitivity to fine-grained, disease-specific features. Together, these results not only reaffirm the model’s superior clustering capacity but also highlight its promise for downstream tasks such as disease stratification, clinical information retrieval, and diagnostic support.

\section{Code, Data, and Model Availability}
Our MedTEB benchmark is publicly available at \url{https://github.com/MohammadKhodadad/MedTEB}, and the MedTE codebase can be found at \url{https://github.com/MohammadKhodadad/MedTE}.  Additionally, the trained MedTE model (CL15, step 8000) is hosted on Hugging Face and can be accessed at \url{https://huggingface.co/MohammadKhodadad/MedTE-cl15-step-8000}.

\section{Conclusion}

We introduced \textbf{MedTE}, a novel medical‐text embedding model trained exclusively via self‐supervised contrastive learning on a richly diverse corpus (PubMed abstracts, Wikipedia medical articles, ClinicalTrials.gov entries, MIMIC-IV notes, bioRxiv/medRxiv preprints, and large-scale medical QA).  We also presented \textbf{MedTEB}, the first large-scale benchmark dedicated to medical embeddings, comprising 51 tasks spanning classification, clustering, pair classification, and retrieval.  Across this suite, MedTE consistently outperforms both domain-specific baselines (e.g., BioBERT, ClinicalBERT, Med-BERT) and leading general-purpose models (e.g., E5, SBERT), achieving an average score of 0.578 versus 0.525 for the nearest competitor, and establishing a new state of the art in medical text representation.

\subsection{Challenges}

Despite these advances, the variability and complexity of medical language demand even broader, more diverse training data, especially for low-resource specialties and rare diseases, to capture nuanced semantic relationships.  Moreover, domain shifts between sources (e.g., clinical notes versus scientific abstracts) can still degrade performance on niche tasks, highlighting the need for more finely granulated corpora and adaptive sampling strategies.

\subsection{Future Work}

Possible extensions could include: first, the development of a dedicated \emph{reranker} trained on medical QA pairs to boost retrieval precision in downstream applications; and second, the integration of {MedTE} alongside this {reranker} within an end‑to‑end QA-RAG pipeline augmenting an {Agentic AI}, thereby enabling dynamic decisions about when and how to retrieve, rerank, and respond to complex medical queries in real time. Together, these directions could push the frontier of intelligent medical NLP toward more accurate, context‑aware clinical decision‑support systems.

\section{Acknowledgements}   
This work was supported by \textbf{MITACS} (grant IT32409). We are also grateful to  \textbf{Pankaj Singh} for his invaluable insights into working with medical data.

\vspace*{ 1 cm}

\bibliographystyle{unsrt}  
\bibliography{references}

\begin{thebibliography}{10}

\bibitem{khattak2019survey}
Faiza~Khan Khattak, Serena Jeblee, Chlo{\'e} Pou-Prom, Mohamed Abdalla, Christopher Meaney, and Frank Rudzicz.
\newblock A survey of word embeddings for clinical text.
\newblock {\em Journal of Biomedical Informatics}, 100:100057, 2019.

\bibitem{lee2020biobert}
Jinhyuk Lee, Wonjin Yoon, Sungdong Kim, Donghyeon Kim, Sunkyu Kim, Chan~Ho So, and Jaewoo Kang.
\newblock Biobert: a pre-trained biomedical language representation model for biomedical text mining.
\newblock {\em Bioinformatics}, 36(4):1234--1240, 2020.

\bibitem{devlin2018bert}
Jacob Devlin.
\newblock Bert: Pre-training of deep bidirectional transformers for language understanding.
\newblock {\em arXiv preprint arXiv:1810.04805}, 2018.

\bibitem{huang2019clinicalbert}
Kexin Huang, Jaan Altosaar, and Rajesh Ranganath.
\newblock Clinicalbert: Modeling clinical notes and predicting hospital readmission.
\newblock {\em arXiv preprint arXiv:1904.05342}, 2019.

\bibitem{rasmy2021med}
Laila Rasmy, Yang Xiang, Ziqian Xie, Cui Tao, and Degui Zhi.
\newblock Med-bert: pretrained contextualized embeddings on large-scale structured electronic health records for disease prediction.
\newblock {\em NPJ digital medicine}, 4(1):86, 2021.

\bibitem{wang2022text}
Liang Wang, Nan Yang, Xiaolong Huang, Binxing Jiao, Linjun Yang, Daxin Jiang, Rangan Majumder, and Furu Wei.
\newblock Text embeddings by weakly-supervised contrastive pre-training.
\newblock {\em arXiv preprint arXiv:2212.03533}, 2022.

\bibitem{reimers2019sentence}
N~Reimers.
\newblock Sentence-bert: Sentence embeddings using siamese bert-networks.
\newblock {\em arXiv preprint arXiv:1908.10084}, 2019.

\bibitem{muennighoff2022mteb}
Niklas Muennighoff, Nouamane Tazi, Lo{\"\i}c Magne, and Nils Reimers.
\newblock Mteb: Massive text embedding benchmark.
\newblock {\em arXiv preprint arXiv:2210.07316}, 2022.

\bibitem{bai2018ehr}
Tian Bai, Ashis~Kumar Chanda, Brian~L Egleston, and Slobodan Vucetic.
\newblock Ehr phenotyping via jointly embedding medical concepts and words into a unified vector space.
\newblock {\em BMC medical informatics and decision making}, 18(Suppl 4):123, 2018.

\bibitem{glicksberg2018automated}
Benjamin~S Glicksberg, Riccardo Miotto, Kipp~W Johnson, Khader Shameer, Li~Li, Rong Chen, and Joel~T Dudley.
\newblock Automated disease cohort selection using word embeddings from electronic health records.
\newblock In {\em PACIFIC SYMPOSIUM on BIOCOMPUTING 2018: Proceedings of the Pacific Symposium}, pages 145--156. World Scientific, 2018.

\bibitem{lewis2020retrieval}
Patrick Lewis, Ethan Perez, Aleksandra Piktus, Fabio Petroni, Vladimir Karpukhin, Naman Goyal, Heinrich K{\"u}ttler, Mike Lewis, Wen-tau Yih, Tim Rockt{\"a}schel, et~al.
\newblock Retrieval-augmented generation for knowledge-intensive nlp tasks.
\newblock {\em Advances in neural information processing systems}, 33:9459--9474, 2020.

\bibitem{yeganova2020better}
Lana Yeganova, Sun Kim, Qingyu Chen, Grigory Balasanov, W~John Wilbur, and Zhiyong Lu.
\newblock Better synonyms for enriching biomedical search.
\newblock {\em Journal of the American Medical Informatics Association}, 27(12):1894--1902, 2020.

\bibitem{zhang2020combining}
Dongdong Zhang, Changchang Yin, Jucheng Zeng, Xiaohui Yuan, and Ping Zhang.
\newblock Combining structured and unstructured data for predictive models: a deep learning approach.
\newblock {\em BMC medical informatics and decision making}, 20(1):280, 2020.

\bibitem{beltagy2019scibert}
Iz~Beltagy, Kyle Lo, and Arman Cohan.
\newblock Scibert: A pretrained language model for scientific text.
\newblock {\em arXiv preprint arXiv:1903.10676}, 2019.

\bibitem{Chen2024M3Embed}
Jianlv Chen, Shitao Xiao, Peitian Zhang, Kun Luo, Defu Lian, and Zheng Liu.
\newblock {M3\textendash Embedding}: Multi{\textendash}linguality, multi{\textendash}functionality, multi{\textendash}granularity text embeddings through self{\textendash}knowledge distillation.
\newblock {\em arXiv preprint}, arXiv:2402.03216, 2024.

\bibitem{li2023towards}
Zehan Li, Xin Zhang, Yanzhao Zhang, Dingkun Long, Pengjun Xie, and Meishan Zhang.
\newblock Towards general text embeddings with multi-stage contrastive learning.
\newblock {\em arXiv preprint arXiv:2308.03281}, 2023.

\bibitem{nussbaum2024nomic}
Zach Nussbaum, John~X Morris, Brandon Duderstadt, and Andriy Mulyar.
\newblock Nomic embed: Training a reproducible long context text embedder.
\newblock {\em arXiv preprint arXiv:2402.01613}, 2024.

\bibitem{tai2020exbert}
Wen Tai, HT~Kung, Xin~Luna Dong, Marcus Comiter, and Chang-Fu Kuo.
\newblock exbert: Extending pre-trained models with domain-specific vocabulary under constrained training resources.
\newblock In {\em Findings of the Association for Computational Linguistics: EMNLP 2020}, pages 1433--1439, 2020.

\bibitem{Yang2022GatorTron}
Xi~Yang, Aokun Chen, Nima PourNejatian, Hoo~Chang Shin, Kaleb~E. Smith, et~al.
\newblock Gatortron: A large clinical language model to unlock patient information from unstructured electronic health records.
\newblock {\em arXiv preprint}, arXiv:2203.03540, 2022.

\bibitem{kanakarajan2022biosimcse}
Kamal~Raj Kanakarajan, Bhuvana Kundumani, Abhijith Abraham, and Malaikannan Sankarasubbu.
\newblock Biosimcse: Biomedical sentence embeddings using contrastive learning.
\newblock In {\em Proceedings of the 13th International Workshop on Health Text Mining and Information Analysis (LOUHI)}, pages 81--86, 2022.

\bibitem{abro2024self}
Waheed~Ahmed Abro, Hanane Kteich, and Zied Bouraoui.
\newblock Self-supervised segment contrastive learning for medical document representation.
\newblock In {\em International Conference on Artificial Intelligence in Medicine}, pages 312--321. Springer, 2024.

\bibitem{min2024adaption}
Lingtong Min, Ziman Fan, Feiyang Dou, Jiaao Sun, Changsheng Luo, and Qinyi Lv.
\newblock Adaption bert for medical information processing with chatgpt and contrastive learning.
\newblock {\em Electronics}, 13(13):2431, 2024.

\bibitem{kailas2023notecontrast}
Prajwal Kailas, Max Homilius, Rahul~C Deo, and Calum~A MacRae.
\newblock Notecontrast: Contrastive language-diagnostic pretraining for medical text.
\newblock In {\em Machine Learning for Health (ML4H)}, pages 201--216. PMLR, 2023.

\bibitem{zhang2022metadata}
Yu~Zhang, Zhihong Shen, Chieh-Han Wu, Boya Xie, Junheng Hao, Ye-Yi Wang, Kuansan Wang, and Jiawei Han.
\newblock Metadata-induced contrastive learning for zero-shot multi-label text classification.
\newblock In {\em Proceedings of the ACM Web Conference 2022}, pages 3162--3173, 2022.

\bibitem{balachandran2024medembed}
Abhinand Balachandran.
\newblock Medembed: Medical-focused embedding models, 2024.

\bibitem{Remy2023BioLORD}
Fran{\c{c}}ois Remy, Kris Demuynck, and Thomas Demeester.
\newblock {BioLORD{\textendash}2023}: Semantic textual representations fusing large language models and clinical knowledge graph insights.
\newblock {\em Journal of the American Medical Informatics Association}, 31(9):1844--1855, 2024.

\bibitem{kasmaee2024chemteb}
Ali~Shiraee Kasmaee, Mohammad Khodadad, Mohammad~Arshi Saloot, Nick Sherck, Stephen Dokas, Hamidreza Mahyar, and Soheila Samiee.
\newblock Chemteb: Chemical text embedding benchmark, an overview of embedding models performance \& efficiency on a specific domain.
\newblock {\em arXiv preprint arXiv:2412.00532}, 2024.

\bibitem{soffer2024scalable}
Shelly Soffer, Benjamin~S Glicksberg, Patricia Kovatch, Orly Efros, Robert Freeman, Alexander~W Charney, Girish~N Nadkarni, and Eyal Klang.
\newblock A scalable framework for benchmarking embedding models for semantic medical tasks.
\newblock {\em medRxiv}, pages 2024--08, 2024.

\bibitem{hemteb2024}
Clinia.
\newblock Introducing hemteb: An open-source benchmark for health information retrieval, 2024.

\bibitem{Lahiri2024TerminologyBenchmark}
Aditya Lahiri, Sangeeta Shukla, Ben Stear, and Taha Mohseni~Ahooyi.
\newblock Benchmarking transformer embedding models for biomedical terminology standardization.
\newblock {\em Machine Learning with Applications}, 21:100413, 2025.

\bibitem{pubmed-site}
{National Library of Medicine (US)}.
\newblock Pubmed, 1996.
\newblock Updated 30 May 2025; accessed 28 Jun 2025.

\bibitem{pmc-site}
{National Library of Medicine (US)}.
\newblock Pubmed central, 2000.
\newblock Updated 12 Jun 2025; accessed 28 Jun 2025.

\bibitem{johnson2023mimic}
Alistair~EW Johnson, Lucas Bulgarelli, Lu~Shen, Alvin Gayles, Ayad Shammout, Steven Horng, Tom~J Pollard, Sicheng Hao, Benjamin Moody, Brian Gow, et~al.
\newblock Mimic-iv, a freely accessible electronic health record dataset.
\newblock {\em Scientific data}, 10(1):1, 2023.

\bibitem{ctgov-site}
{National Library of Medicine (US)}.
\newblock Clinicaltrials.gov, 2000.
\newblock Updated 18 Jun 2025; accessed 28 Jun 2025.

\bibitem{biorxiv-2019}
Richard Sever, Ted Roeder, Samantha Hindle, and et~al.
\newblock biorxiv: the preprint server for biology.
\newblock {\em bioRxiv}, 2019.
\newblock preprint.

\bibitem{medrxiv-site}
medRxiv.
\newblock medrxiv, 2019.
\newblock accessed 28 Jun 2025.

\bibitem{pal2022medmcqa}
Ankit Pal, Logesh~Kumar Umapathi, and Malaikannan Sankarasubbu.
\newblock Medmcqa: A large-scale multi-subject multi-choice dataset for medical domain question answering.
\newblock In {\em Conference on health, inference, and learning}, pages 248--260. PMLR, 2022.

\bibitem{jin2021disease}
Di~Jin, Eileen Pan, Nassim Oufattole, Wei-Hung Weng, Hanyi Fang, and Peter Szolovits.
\newblock What disease does this patient have? a large-scale open domain question answering dataset from medical exams.
\newblock {\em Applied Sciences}, 11(14):6421, 2021.

\bibitem{BenAbacha-BMC-2019}
Asma {Ben Abacha} and Dina Demner{-}Fushman.
\newblock A question-entailment approach to question answering.
\newblock {\em {BMC} Bioinform.}, 20(1):511:1--511:23, 2019.

\bibitem{voorhees2021trec}
Ellen Voorhees, Tasmeer Alam, Steven Bedrick, Dina Demner-Fushman, William~R Hersh, Kyle Lo, Kirk Roberts, Ian Soboroff, and Lucy~Lu Wang.
\newblock Trec-covid: constructing a pandemic information retrieval test collection.
\newblock In {\em ACM SIGIR Forum}, volume~54, pages 1--12. ACM New York, NY, USA, 2021.

\bibitem{boteva2016}
Vera Boteva, Demian Gholipour, Artem Sokolov, and Stefan Riezler.
\newblock A full-text learning to rank dataset for medical information retrieval.
\newblock 2016.

\bibitem{loshchilov2017decoupled}
Ilya Loshchilov and Frank Hutter.
\newblock Decoupled weight decay regularization.
\newblock {\em arXiv preprint arXiv:1711.05101}, 2017.

\bibitem{rasley2020deepspeed}
Jeff Rasley, Samyam Rajbhandari, Olatunji Ruwase, and Yuxiong He.
\newblock Deepspeed: System optimizations enable training deep learning models with over 100 billion parameters.
\newblock In {\em Proceedings of the 26th ACM SIGKDD international conference on knowledge discovery \& data mining}, pages 3505--3506, 2020.

\end{thebibliography}

\appendix

\subsection{wiki-pubmed and pubmed-mimic-IV}
\begin{figure*}[!t]
  \centering
  \subfloat[PubMed vs MIMIC-IV]{
    \includegraphics[width=0.48\textwidth]{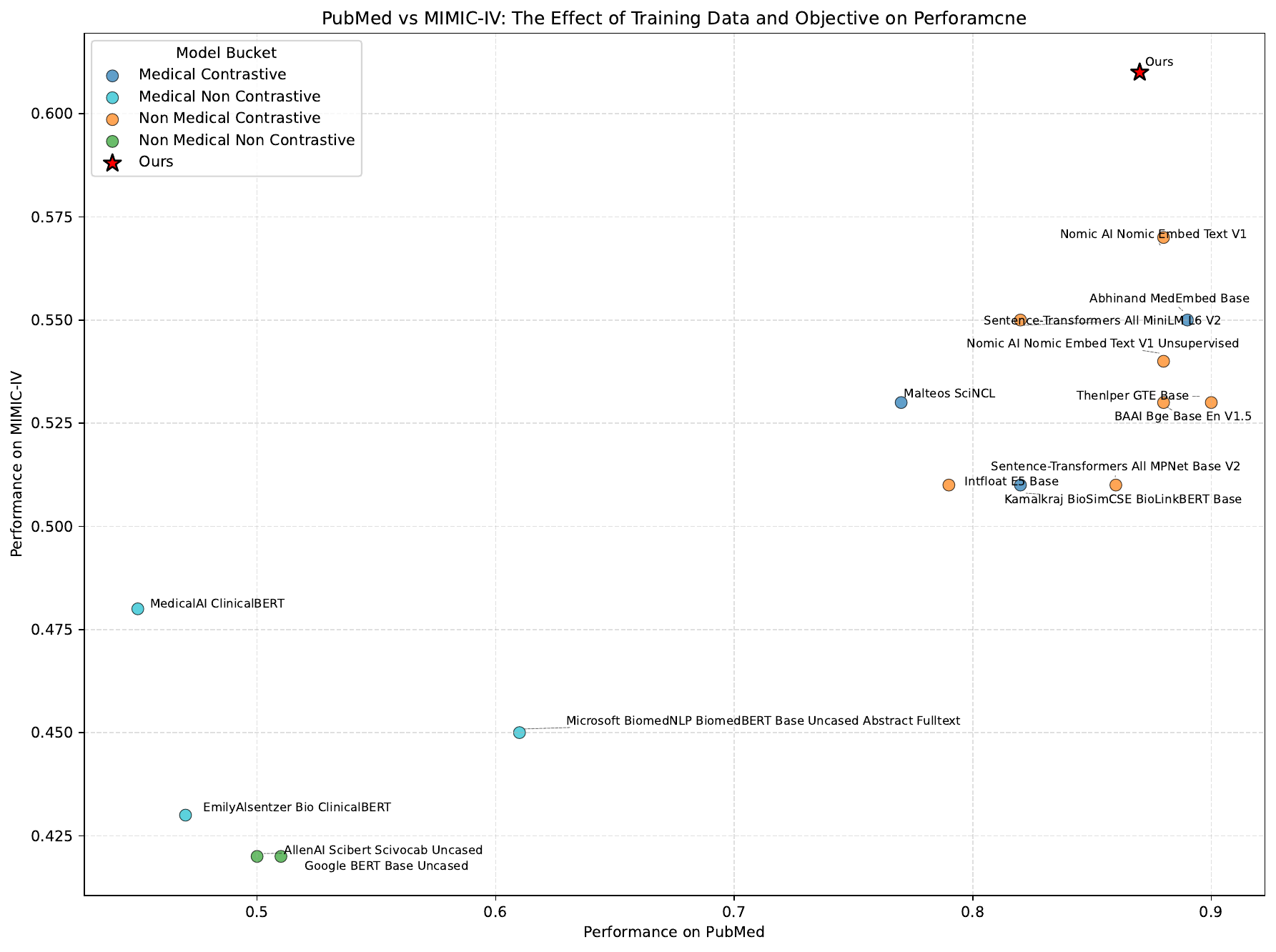}
    \label{fig:pubmed_vs_mimc}
  }
  \subfloat[PubMed vs Wikipedia]{
    \includegraphics[width=0.48\textwidth]{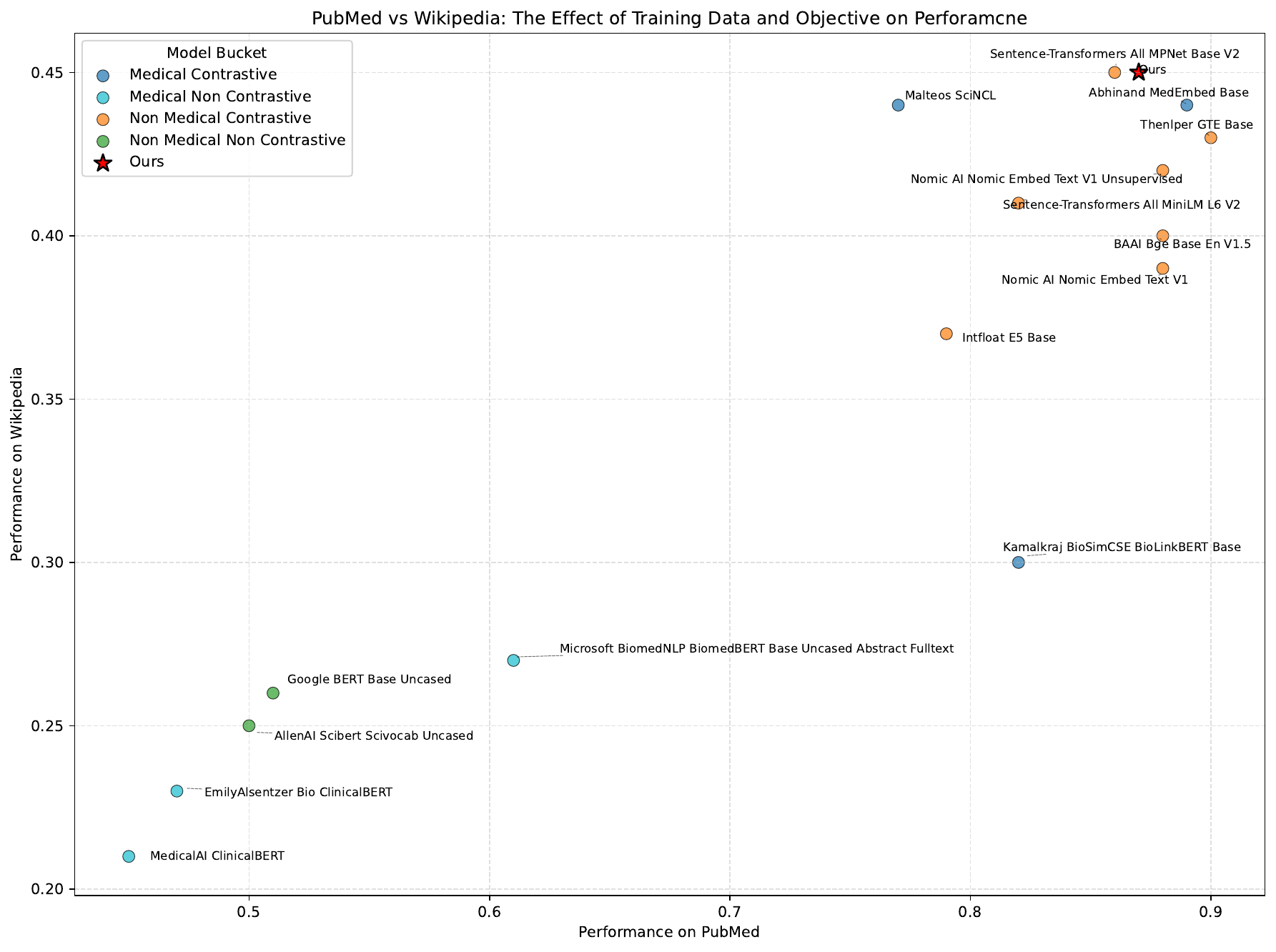}
    \label{fig:pubmed_vs_wiki}
  }
  \caption{Comparison of Models' performance on different sources of tasks in MedTEB}
  \label{fig:style_comparison}
\end{figure*}

In Figure \ref{fig:wiki_vs_mimc}, we compare the performance of all evaluated models on the MIMIC-IV and Wikipedia datasets, revealing a clear gap between clinical-style and general-domain text. Figures \ref{fig:pubmed_vs_mimc} and \ref{fig:pubmed_vs_wiki} extend this analysis by incorporating PubMed, a large, peer-reviewed biomedical corpus, alongside MIMIC-IV and Wikipedia. By placing PubMed in these pairwise comparisons, we can directly assess how exposure to formal biomedical language affects each model’s ability to generalize across domains. In particular, models that leverage domain-specific pretraining on PubMed tend to narrow the performance gap on MIMIC-IV, while still outperforming on Wikipedia, underscoring both the benefits and limits of specialized corpora when applied to heterogeneous biomedical and general-knowledge text.



\begin{figure*}[htbp]
  \centering

  \subfloat[][Evaluation time vs.\ effectiveness across all task categories
              \label{fig:perf_time_all}]{
    \includegraphics[width=0.48\linewidth]{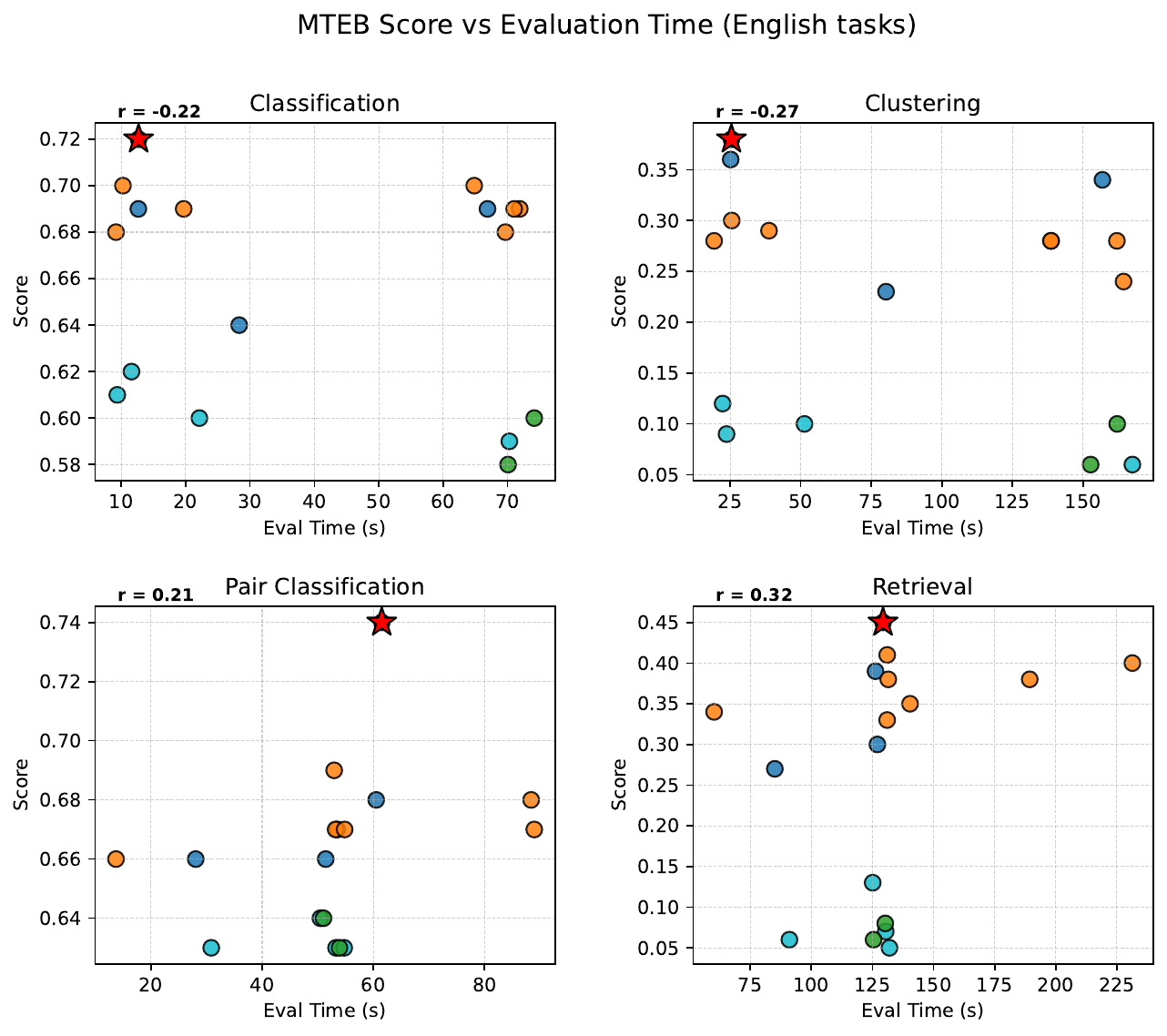}
  }
  \subfloat[][Per-model average performance and evaluation time
              \label{fig:perf_time_avg}]{
    \includegraphics[width=0.48\linewidth]{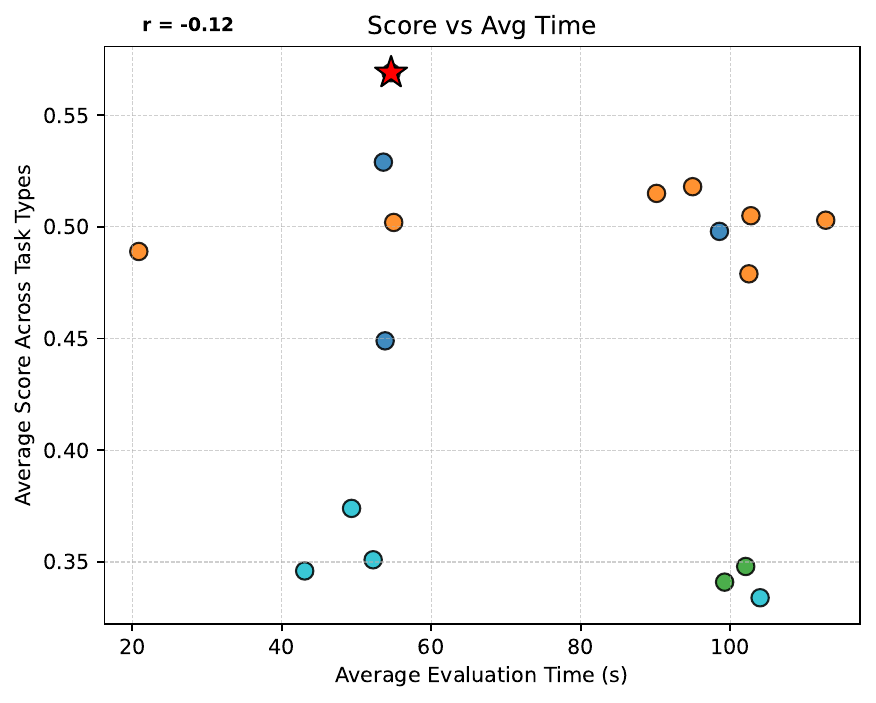}
  }

  \vspace{1em}

  \subfloat[][Legend for the four model groups
              \label{fig:perf_time_legend}]{
    \includegraphics[width=0.30\linewidth]{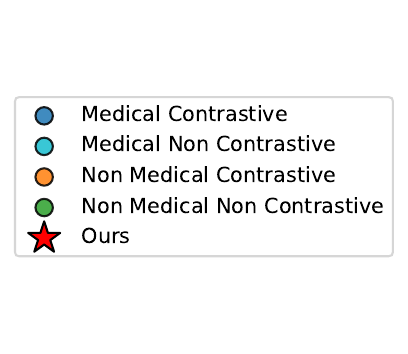}
  }

  \caption{%
    Visualization of Average Score vs Time
  }
  \label{fig:perf_time_combined}
\end{figure*}

\vspace{2cm}
\subsection{Loss}
\begin{figure*}[htpb]
  \centering
  \includegraphics[width=\textwidth]{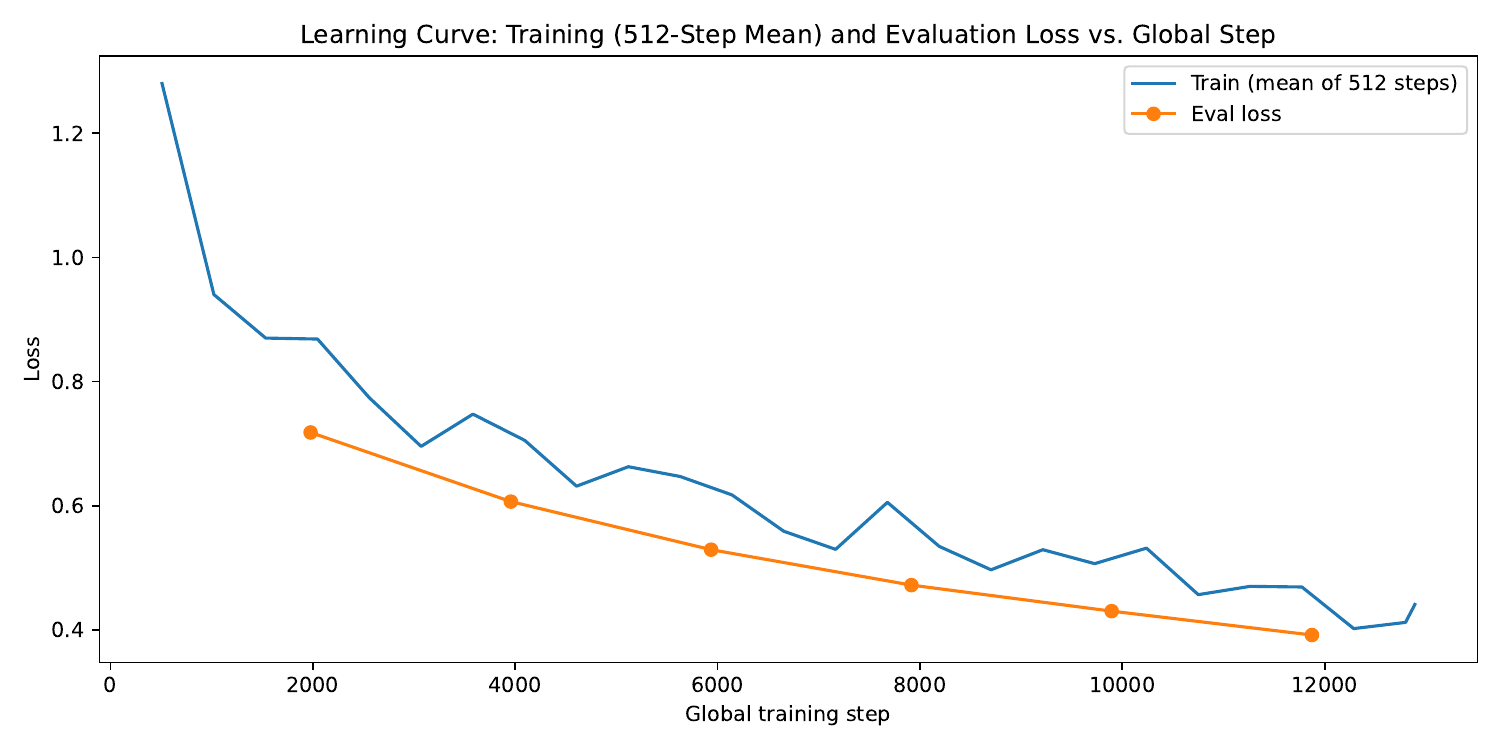}
  \caption{Training and validation loss curves for
  \textsc{MedTE}.}
  \label{fig:loss_curve}
\end{figure*}

Figure~\ref{fig:loss_curve} shows the training and validation loss over our entire pre‑training schedule. Although both curves steadily decrease, the average MedTEB score peaks at around step 6000 and then declines by the time it reaches step 8000. This discrepancy arises because reductions in contrastive loss do not always translate into stronger downstream performance, especially when benchmark data is kept separate from both training and validation sets. In other words, even as contrastive loss improves, downstream evaluation remains essential to verify real gains.

\subsection{Runtime–effectiveness overview}
Figure~\ref{fig:perf_time_all} contrasts \emph{evaluation time} (seconds) with \emph{task‐specific effectiveness} for the four benchmark families, including classification, clustering, pair classification, and retrieval.  In every scatter our model appears in the upper-left region, reflecting both high accuracy and low latency.  Pearson correlations range from \(r=-0.32\) (retrieval) to \(r=-0.07\) (clustering), indicating only a weak and sometimes negative association between longer evaluation times and higher scores.  Figure~\ref{fig:perf_time_avg} also shows the same data into a per-model average, yielding a similarly modest overall trend (\(r=-0.12\)) and confirming that efficient contrastive training can achieve state-of-the-art performance without incurring additional inference cost.

\end{document}